\documentclass{article}

\usepackage{arxiv}

\usepackage[utf8]{inputenc} 
\usepackage[T1]{fontenc}    
\usepackage{hyperref}       
\usepackage{url}            
\usepackage{booktabs}       
\usepackage{amsmath,amssymb,amsfonts}       
\usepackage{nicefrac}       
\usepackage{microtype}      
\usepackage{lipsum}
\usepackage{graphicx}
\usepackage{abstract}
\graphicspath{ {./images/} }

\title{Spectral Graph Neural Networks for Cognitive Task Classification in fMRI Connectomes}

\author{
 Debasis Maji \\
  Department of Computer \& System Sciences\\
  Visva-Bharati\\
  India, 731235 \\
  \texttt{youdebasis@gmail.com} \\
  \AND
 Arghya Banerjee \\
  Department of Computer Science \& Engineering(AIML)\\
  Institute of Engineering \& Management\\
  India, 700091 \\
  \texttt{banerjeearghya2004@gmail.com} \\
   \And
 Debaditya Barman \\
  Department of Computer \& System Sciences\\
  Visva-Bharati\\
  India, 731235 \\
  \texttt{debadityabarman@gmail.com} \\
}

\begin{document}
\maketitle
\begin{abstract}
Cognitive task classification using machine learning plays a central role in decoding brain states from neuroimaging data. By integrating machine learning with brain network analysis, complex connectivity patterns can be extracted from functional magnetic resonance imaging connectomes. This process transforms raw blood-oxygen-level-dependent (BOLD) signals into interpretable representations of cognitive processes. Graph neural networks (GNNs) further advance this paradigm by modeling brain regions as nodes and functional connections as edges, capturing topological dependencies and multi-scale interactions that are often missed by conventional approaches. Our proposed SpectralBrainGNN model, a spectral convolution framework based on graph Fourier transforms (GFT) computed via normalized Laplacian eigendecomposition. Experiments on the Human Connectome Project-Task (HCPTask) dataset demonstrate the effectiveness of the proposed approach, achieving a classification accuracy of 96.25\%. The implementation is publicly available at \url{https://github.com/gnnplayground/SpectralBrainGNN} to support reproducibility and future research.
\end{abstract}


\section{Introduction}
One of the central objectives of modern multidisciplinary science is to figure out how the human brain works. A lot of research is currently focused on figuring out how to decode complex patterns in brain activity to help with things like figuring out cognitive processes and diagnosing mental health disorders\cite{jo2019deep,eslami2019asd}. This research heavily depends on neuroimaging techniques including electroencephalography (EEG), magnetoencephalography (MEG), and functional magnetic resonance imaging (fMRI). These techniques are used to map brain activity and identify the functional networks that support various cognitive functions\cite{fox2007spontaneous,zhang2024metarlec}. Among these fMRI has emerged as a potential candidate due to its unique spatial and temporal resolution, measuring blood-oxygen-level-dependent (BOLD) responses that indicate changes in neural activity\cite{kohoutova2020toward,davis2020discovery}. This capability allows researchers to explore cognitive complexities and identify biomarkers for neurological disorders\cite{bassett2017network,yang2022data}. BOLD signals are commonly used to construct functional connectivity (FC) networks from fMRI data, in which brain regions are connected by correlations associated with either normal or pathological conditions\cite{kawahara2017brainnetcnn,morris2019weisfeiler}. \par
The choice of data representation significantly impacts brain-state classification outcomes across modalities\cite{craik2019deep,saranskaia2025aim,vaghari2022late}, which results in the utilization of network-based approaches to capture both local and global connectivity patterns. \cite{wang2010graph,richiardi2011decoding,takerkart2014graph}. To model these FC features effectively, brain networks are typically represented as graphs, with nodes as regions of interest (ROIs) and edges as functional connections, often computed via metrics like Pearson's correlation coefficients\cite{cui2022positional,said2023neurograph,saeidi2022decoding,gorban2021dynamic,li2021braingnn}. \par 
This graph architecture facilitates the explicit representation of inter-region correlations. This allows the implementation of graph machine learning (GML) techniques, particularly graph neural networks (GNNs), which incorporate local information to identify patterns in functional connectivity\cite{wang2022graph,li2021braingnn,saeidi2022decoding}. Recent advancements include ensemble learning for constructing ``synolitic" graphs that combine multiple connectivity metrics to handle noise, inter-subject variability, and multimodal data, reducing computational demands while enhancing classification accuracy\cite{vlasenko2024ensemble,mohammed2023comprehensive}. However, existing GNN-based approaches\cite{li2021braingnn,vlasenko2024ensemble} often overlook inherent activation pathways in brain networks—sequences of neural signal transmissions that form paths across functional connectivity (FC) graphs \cite{sankar2018dynamic,keller2018task}. These pathways capture how brain regions organize into functional modules and coordinate during tasks and ignoring them may miss complex interactions beyond pairwise correlations.

\par Inspired by this, we propose SpectralBrainGNN, a spectral graph neural network that uses exact graph Fourier transforms (GFT) via Laplacian eigendecomposition to operate in the frequency domain. Unlike spatial GNNs that aggregate local neighborhoods, SpectralBrainGNN captures frequency-specific motifs, that provides sharper selectivity and improved interpretability for fMRI-derived connectomes.

Our contributions in this paper are as follows:
\begin{itemize}
\item A novel exact spectral convolution layer with learnable frequency filters and attention readout for graph-level predictions.
\item Empirical validation on HCP-derived dataset, achieving superior performance (e.g., 96.25\% accuracy on HCPTask) over modern baselines.
\end{itemize}
\section{Related Works}
Brain network analysis, which consists of complex connectivity patterns within the brain, originates from the initial studies of motor pathways, demonstrated by Keller et al.\cite{keller2018task}, who illustrated task-dependent activation of quick and slow motor routes during imagery, reflecting actual movements and establishing a foundation for pathway-specific modeling. In 2020, predictive applications appeared, with He et al.\cite{he2020deep} enhancing connectome-based forecasting of cognitive actions through graph convolutional networks, focusing individual diversity in functional connections. The field accelerated in 2021 with Li et al.'s BrainGNN\cite{li2021braingnn}, an interpretable ROI-aware framework for fMRI that fused local and global features to enhance disorder detection.
\par In the following year, the field gained momentum through several influential contributions. These included BrainGB by Cui et al.\cite{cui2022braingb}, which standardized GNN evaluation benchmarks; the transformer-based modeling of global node interactions proposed by Kan et al.\cite{kan2022brain}; sparsity-guided connectivity analysis for mild cognitive impairment biomarkers introduced by Zhang et al.\cite{zhang2022probing}; adversarial learning for dynamic brain networks by Yang et al.\cite{yang2022data}; self-supervised pre-training strategies developed by Li et al.\cite{li2022brain}; and novel spatiotemporal GNN architectures proposed by Wang, Chen, and Li et al. \cite{wang2022graph}. Together, these advances substantially expanded the application of GNNs in disease identification and cognitive mapping.
\par More recently, research attention shifted toward incorporating node attributes and data augmentation strategies. Representative works include temporal pattern detection for anomaly identification by Said et al.\cite{said2023neurograph} and data-augmented analysis for brain disorder classification by Liu et al.\cite{liu2023coupling}, both of which improved robustness in sparse data settings. More recent studies in 2024 further advanced GNNs by emphasizing interpretability and topological structure, such as semantic bridging between brain and machine representations by Chen et al.\cite{chen2024bridging}, permutation-invariant encoders for neuronal circuits proposed by Liao, Wan, and Du \cite{liao2024joint}, and logits-constrained attention mechanisms for accurate age and intelligence prediction introduced by Xu et al.\cite{xu2024learning}.
Unlike spatial GNNs that aggregate local features in the vertex domain and risk over smoothing, SpectralBrainGNN uses exact graph Fourier transforms with learnable frequency filters to capture global frequency patterns and improve cognitive task classification.

\section{Proposed method}
\subsection{Brain Image to Graph formulation}
For brain fMRI image to graph formulation, the preprocessing and graph construction pipeline proposed by Said et al.\cite{said2023neurograph} has been adopted for this paper. fMRI provides a sequence of volumetric images that capture blood-oxygen-level-dependent (BOLD) signals over time. Mathematically, an fMRI scan for a given subject can be represented as a four-dimensional tensor
\[
\mathcal{F} \in \mathbb{R}^{X \times Y \times Z \times T},
\]
where $(X,Y,Z)$ denote the spatial dimensions of the brain volume and $T$ denotes the number of acquired time points during the scanning session. Each volumetric image encodes the intensity of the BOLD signal at different spatial locations in the brain. Direct analysis of $\mathcal{F}$ is challenging due to its high dimensionality, noise, and strong inter-voxel correlations. Moreover, brain activity is known to exhibit coherent patterns across spatially distributed regions rather than at the level of individual voxels. So the BOLD signal is commonly summarized into a collection of spatially extended functional units, referred to as brain parcels or regions of interest (ROIs). Each ROI consists of a group of voxels whose time-series exhibit temporally correlated activity.
\par To obtain these ROIs, data from the Human Connectome Project (HCP), a publicly available neuroimaging dataset\footnote{https://www.humanconnectome.org/study/hcp-young-adult} containing high-quality fMRI recordings, has been utilized for benchmarking. The group-level Schaefer atlas has been employed to parcellate the cerebral cortex into a predefined number of ROIs. This atlas provides a hierarchical organization of cortical regions at multiple spatial resolutions which supports consistent and reproducible region definitions across subjects.
\par Brain connectivity networks are then constructed to provide a compact and structured representation of functional interactions. In this graph-based abstraction, ROIs are modeled as nodes, while statistical dependencies between ROI time-series are modeled as edges. Such representations have been shown to effectively capture functional relationships relevant to neurological and neurodevelopmental disorders.
Let the set of ROI coordinates defined by the Schaefer atlas be denoted as
\[
\mathcal{C} = \{ c_1, c_2, \dots, c_N \}, \quad c_i \in \mathbb{R}^3,
\]
where $N$ is the number of regions. Around each coordinate $c_i$, a spherical mask of radius $r=5$ mm is defined. The BOLD signals are then extracted from these spheres using a masker function, which returns the mean time-series for each ROI. Thus, each node $v_i \in V$ is associated with a time-series
\[
x_i = (x_i^1, x_i^2, \dots, x_i^T), \quad x_i \in \mathbb{R}^T.
\]
Functional connectivity between pairs of ROIs is quantified using correlation analysis. Given two ROI time-series $x_i$ and $x_j$, their Pearson correlation coefficient is computed as
\[
\rho_{ij} = \frac{\text{Cov}(x_i, x_j)}{\sigma(x_i) \, \sigma(x_j)},
\]
where $\text{Cov}(\cdot,\cdot)$ denotes covariance and $\sigma(\cdot)$ the standard deviation.  

Collecting all pairwise correlations yields a symmetric connectivity matrix
\[
C \in \mathbb{R}^{N \times N}, \quad C_{ij} = \rho_{ij}.
\]
To construct the graph topology, a threshold $\tau$ is applied such that an edge $(v_i, v_j) \in E$ exists if $C_{ij} > \tau$. This thresholding operation enforces sparsity while preserving the strongest functional connections.
Finally, each subject is represented as a weighted, undirected graph
\[
G = (V, E, X),
\]
where $V$ denotes the set of ROIs, $E$ the functional connections, and $X \in \mathbb{R}^{N \times T}$ the node features derived from the BOLD time-series. Each graph is paired with a subject-level label $y \in \{0,1,\cdots\}$, indicating diagnostic group.
\subsection{Spectral Graph Neural Network (SpectralBrainGNN)}
\begin{figure}[ht!]
    \includegraphics[width=\textwidth]{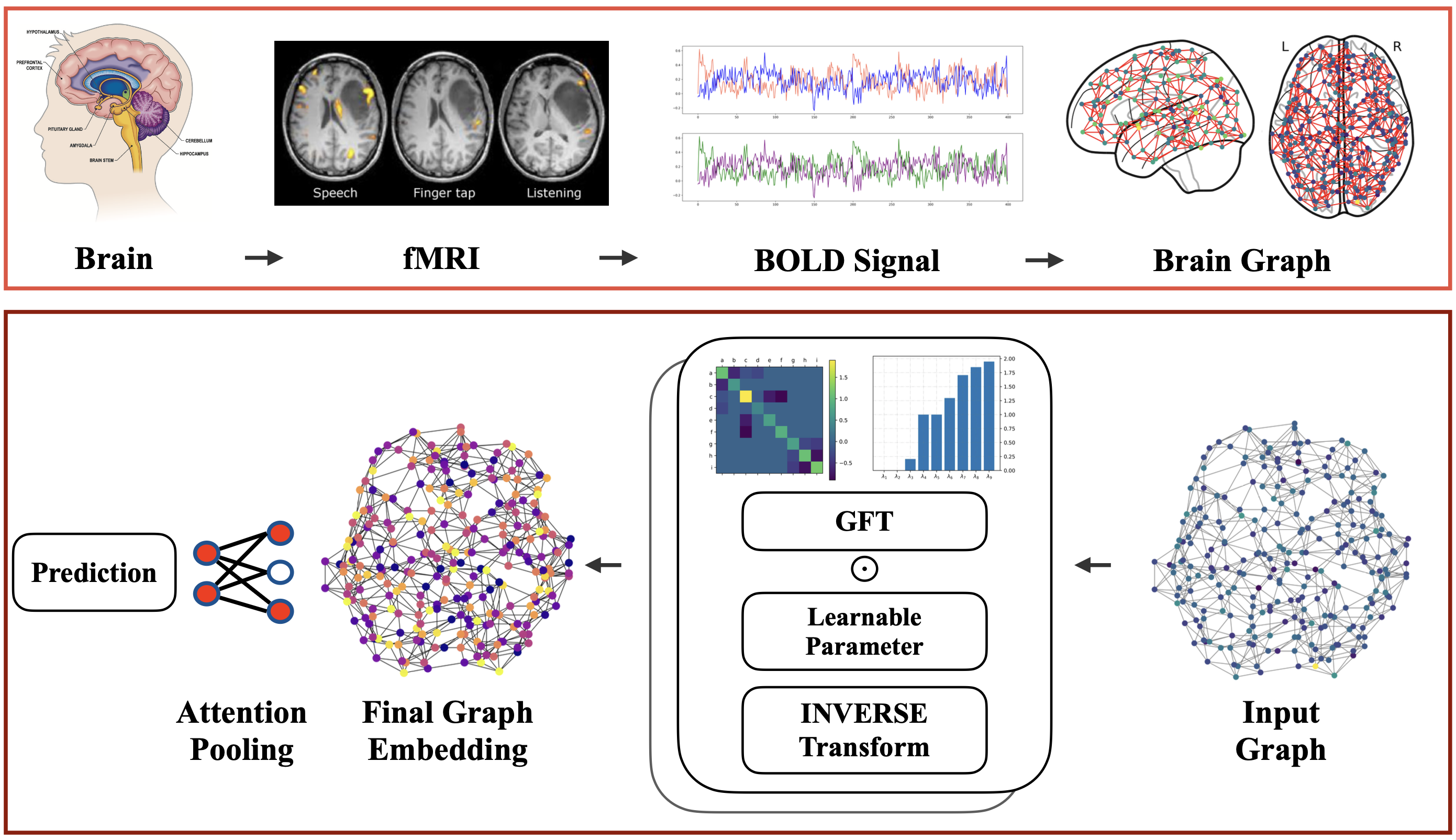}
    \caption{Schematic diagram of SpectralBrainGNN model for Cognitive Task Classification.}
    \label{fig:proposed_model}
\end{figure}

After constructing subject-specific brain graphs $G = (V, E, X)$, we employ a spectral graph neural network for brain image (SpectralBrainGNN) as illustrated in Fig.~\ref{fig:proposed_model},  to learn discriminative representations for classification. Our proposed model, SpectralBrainGNN, consists of $L$ layers. Each layer takes the node embeddings generated by the previous layer as input and outputs the updated node embeddings, progressively refining the representation through spectral-domain filtering at each layer. The method begins with computing the normalized Laplacian of the input graph and proceeds through exact eigendecomposition, graph Fourier transform, learnable frequency filtering, inverse GFT, and channel-wise linear projection.

\subsubsection{Spectral Convolution Layer:}
Given an input graph with node features $H^{(l)} \in \mathbb{R}^{N \times F_{\text{in}}}$ at layer $l$ where $N$ is the number of nodes and $F_{\text{in}}$ is the input feature dimension, we first compute the normalized Laplacian matrix
\begin{equation}
    L = I_N - D^{-1/2} A D^{-1/2},
    \label{eq:laplacian}
\end{equation}
where $A \in \mathbb{R}^{N \times N}$ is the adjacency matrix, $D = \operatorname{diag}(A \mathbf{1})$ is the degree matrix, and $I_N$ is the identity matrix. The symmetric positive semi-definite matrix $L$ admits an eigendecomposition
\begin{equation}
    L = U \Lambda U^T,
\end{equation}
with orthogonal Fourier basis $U \in \mathbb{R}^{N \times N}$ i.e. eigenvectors and eigenvalues $\Lambda = \operatorname{diag}(\lambda_1, \dots, \lambda_N)$ where $0 \leq \lambda_i \leq 2$. For moderate sized graphs, the eigenvectors $U$ and eigenvalues $\lambda$ are computed only once during the initial forward pass and subsequently cached, thereby distributing the $O(N^3)$ computational cost of the eigendecomposition over multiple iterations.

The exact graph Fourier transform projects the node features into the frequency domain:
\begin{equation}
    \hat{H}^{(l)} = U^T H^{(l)} \in \mathbb{R}^{N \times F_{\text{in}}},
    \label{eq:gft}
\end{equation}
where $\hat{H}^{(l)}$ represents the frequency-domain signal. A learnable filter $g(\lambda) \in \mathbb{R}^N$ is then applied element-wise:
\begin{equation}
    \hat{H}^{(l+1)} = g(\lambda) \odot \hat{H}^{(l)},
    \label{eq:filter}
\end{equation}
with $\odot$ denoting Hadamard (element-wise) multiplication broadcast over feature channels. The filter values $g(\lambda_i)$ are parameterized e through a multilayer perceptron $\operatorname{MLP}(\lambda_i; \phi)$ that maps each eigenvalue to a scalar.

The filtered signal is transformed back to the node domain via the inverse GFT:
\begin{equation}
    \tilde{H}^{(l+1)} = U \hat{H}^{(l+1)} \in \mathbb{R}^{N \times F_{\text{in}}},
\end{equation}
followed by a linear channel mixing:
\begin{equation}
    H^{(l+1)} = \tilde{H}^{(l+1)} W + b \in \mathbb{R}^{N \times F_{\text{out}}},
    \label{eq:inv_gft}
\end{equation}
where $W \in \mathbb{R}^{F_{\text{in}} \times F_{\text{out}}}$ and $b \in \mathbb{R}^{F_{\text{out}}}$ are learnable parameters. A ReLU nonlinearity is applied after each convolution layer to introduce non-linearity:
\begin{equation}
    H^{(l+1)} = \operatorname{ReLU}(H^{(l+1)}).
\end{equation}


\subsubsection{Attention-Based Readout:}
After the final convolution layer, the refined node embeddings $H^{(L)} \in \mathbb{R}^{N \times F_h}$ are aggregated into a single graph-level representation using an attention mechanism. Per-node attention scores are computed as
\begin{equation}
    a_i = \sigma \left( H^{(2)}_i \Gamma \right) \in \mathbb{R},
\end{equation}
where $\Gamma \in \mathbb{R}^{F_h}$ is a learnable projection vector and $\sigma$ is the sigmoid function. The scores are normalized to obtain attention weights:
\begin{equation}
    \tilde{a}_i = \frac{a_i}{\sum_{j=1}^N a_j + \epsilon}, \quad \epsilon = 10^{-8}.
\end{equation}
The graph embedding is then formed by the weighted sum:
\begin{equation}
    h_g = \sum_{i=1}^N \tilde{a}_i H^{(2)}_i \in \mathbb{R}^{F_h}.
    \label{eq:readout}
\end{equation}
This adaptive pooling assigns higher importance to salient nodes i.e. functional hubs in brain connectomes, outperforming uniform pooling strategies like mean or max pooling in distinguishing complex graph structures.

\subsubsection{Final Prediction Layer:}
The graph-level embedding $h_g$ is passed through a fully connected layer to produce the final output:
\begin{equation}
    \hat{y} = h_g W_{\text{out}} + b_{\text{out}} \in \mathbb{R}^C,
    \label{eq:predictor}
\end{equation}
where $W_{\text{out}} \in \mathbb{R}^{F_h \times C}$ and $b_{\text{out}} \in \mathbb{R}^C$ are learnable parameters. The entire model is trained end-to-end using cross-entropy loss function for classification.

\section{Result \& Discussion}
\subsection{Implementation Details}
The implementation of the SpectralBrainGNN model involves PyTorch and leverages an Apple M4 chip with 16 GB unified memory to enhance computational efficiency for spectral operations. The SpectralBrainGNN framework comprises 2 spectral convolution modules, each performing exact graph Fourier transforms with learnable frequency filters, followed by an attention-based readout layer. These modules are stacked with ReLU nonlinear activations and incorporated into the framework, which also consists of 1 fully connected layer for final prediction. Both the spectral convolution and fully connected layers employ the ReLU nonlinear activation function. The number of neurons in each spectral convolution layer is specified as 64. For the fully connected layer, we have chosen to use 64 neurons. To address the issue of overfitting, a dropout rate of 0.5 has been implemented for these layers. The optimization process involves utilizing the Adaptive Moment Estimation (Adam) optimizer with a learning rate set at 0.001 and a regularization parameter of 0.0001. The model is trained for 200 epochs, and this training process is repeated for 10 times to achieve better result stability.
\subsection{Evaluation metrics}
Given the limited sample size, we employed random selection to allocate 60\% of samples to training, 20\% to validation, and 20\% to testing. During training, we performed validation at each iteration and saved parameters yielding the highest validation performance, which were then used for testing. This procedure was repeated 30 times, with average results and standard deviations recorded for each method. For fair comparison, identical data splits and training strategies were applied across our SpectralBrainGNN and competing methods. Model efficacy was assessed using five metrics: accuracy, precision, recall and F1-score. These provide a comprehensive performance evaluation, where true positives (TP), true negatives (TN), false positives (FP), and false negatives (FN) denote correctly/erroneously classified positive/negative samples, respectively. The four metrics are defined as:
\begin{align}
\text{Accuracy} &= \frac{\text{TP} + \text{TN}}{\text{TP} + \text{TN} + \text{FP} + \text{FN}}, \\
\text{Precision} &= \frac{\text{TP}}{\text{TP} + \text{FP}}, \\
\text{Recall} &= \frac{\text{TP}}{\text{TP} + \text{FN}},\\
\text{F1-score} &= 2 \cdot \frac{\text{Precision} \times \text{Recall}}{\text{Precision} + \text{Recall}}.
\end{align}
\subsection{Dataset}
The Human Connectome Project (HCP) Young Adult dataset~\cite{van2013wu} forms the basis of our experimental benchmarks and provides high-quality multimodal neuroimaging data from over 1,200 healthy adults aged 22 to 35 years. This dataset is used to model functional connectivity patterns associated with cognitive traits. The task-based classification benchmarks are derived from the NeuroGraph framework~\cite{said2023neurograph}, which supports standardized evaluation of SpectralBrainGNN against baseline graph classification methods.

\par The resulting HCPTask dataset consists of 7,443 graphs constructed from task-evoked fMRI recordings across seven cognitive paradigms: emotion processing, gambling, language comprehension, motor execution, relational processing, social cognition, and working memory~\cite{barch2013function}. Each graph contains 400 regions of interest (ROIs) and is used in a multi-class classification setting to identify the underlying task state from functional connectivity patterns. This experimental setup evaluates the model’s ability to capture dynamic neural activations under balanced class distributions, supporting robust training and evaluation.
\subsection{Classification Result}

\begin{table}[ht!]
\centering
\caption{Classification performance on the HCPTask dataset. Results are reported as mean $\pm$ standard deviation over 30 runs.}
\label{tab:results}
\begin{tabular}{lcccc}
\hline
\textbf{Model} & \textbf{Accuracy (\%)} & \textbf{Precision (\%)} & \textbf{Recall (\%)} & \textbf{F1-Score (\%)} \\
\hline
GCN\cite{kipf2016semi} & 86.29 $\pm$ 0.98 & 85.12 $\pm$ 1.05 & 86.45 $\pm$ 0.92 & 85.78 $\pm$ 0.97 \\
GAT\cite{velivckovic2017graph} & 85.60 $\pm$ 1.26 & 84.78 $\pm$ 1.31 & 85.92 $\pm$ 1.18 & 85.35 $\pm$ 1.24 \\
SAGE\cite{hamilton2017inductive} & 84.49 $\pm$ 0.57 & 83.67 $\pm$ 0.63 & 84.81 $\pm$ 0.55 & 84.24 $\pm$ 0.59 \\
ResGCN\cite{pei2022resgcn} & 93.75 $\pm$ 0.35 & 93.02 $\pm$ 0.41 & 93.89 $\pm$ 0.33 & 93.45 $\pm$ 0.38 \\
GraphGPS\cite{rampavsek2022recipe} & 92.13 $\pm$ 2.00 & 91.45 $\pm$ 2.12 & 92.28 $\pm$ 1.95 & 91.86 $\pm$ 2.04 \\
Graph-Mamba\cite{wang2024graph} & 94.17 $\pm$ 0.86 & 93.56 $\pm$ 0.90 & 94.03 $\pm$ 0.84 & 93.79 $\pm$ 0.88 \\
BrainMAP\cite{wang2025brainmap} & 94.74 $\pm$ 0.07 & 94.12 $\pm$ 0.10 & \textbf{94.68 $\pm$ 0.06} & 94.40 $\pm$ 0.08 \\
\textbf{SpectralBrainGNN} & \textbf{96.25 $\pm$ 1.37} & \textbf{95.46 $\pm$ 1.42} & 94.32 $\pm$ 1.51 & \textbf{95.58 $\pm$ 1.39} \\
\hline
\end{tabular}
\end{table}

The classification results on the HCPTask dataset demonstrate the strong performance of the proposed SpectralBrainGNN in decoding cognitive task states from functional connectomes. As summarized in Table~\ref{tab:results}, SpectralBrainGNN achieves the highest accuracy of 96.25\%, surpassing competitive baselines such as BrainMAP (94.74\%) and Graph-Mamba (94.17\%) by margins of 1.51\% and 2.08\%, respectively. This improvement is consistently reflected in precision (95.46\%) and F1-score (95.58\%), indicating reliable discrimination across the seven task conditions, including emotion and motor processing.

\par Although the recall (94.32\%) is slightly lower than that of BrainMAP (94.68\%), this difference suggests a modest trade-off that favors higher precision for certain classes. In contrast, earlier architectures such as GCN and GAT achieve accuracies around 85–86\%, revealing limitations in modeling frequency-specific connectivity patterns. More recent methods, including ResGCN and GraphGPS, improve representational capacity but remain less effective than the proposed exact Fourier-based filtering approach. The consistently low standard deviations across runs further confirm the stability and robustness of SpectralBrainGNN, highlighting its suitability for neuroimaging classification tasks. A paired t-test have been conducted on the accuracy scores across 30 runs, yielding a p-value of 0.028, showing that the improvement over the best baseline is statistically significant
\begin{figure}[ht!]
    \includegraphics[width=\textwidth]{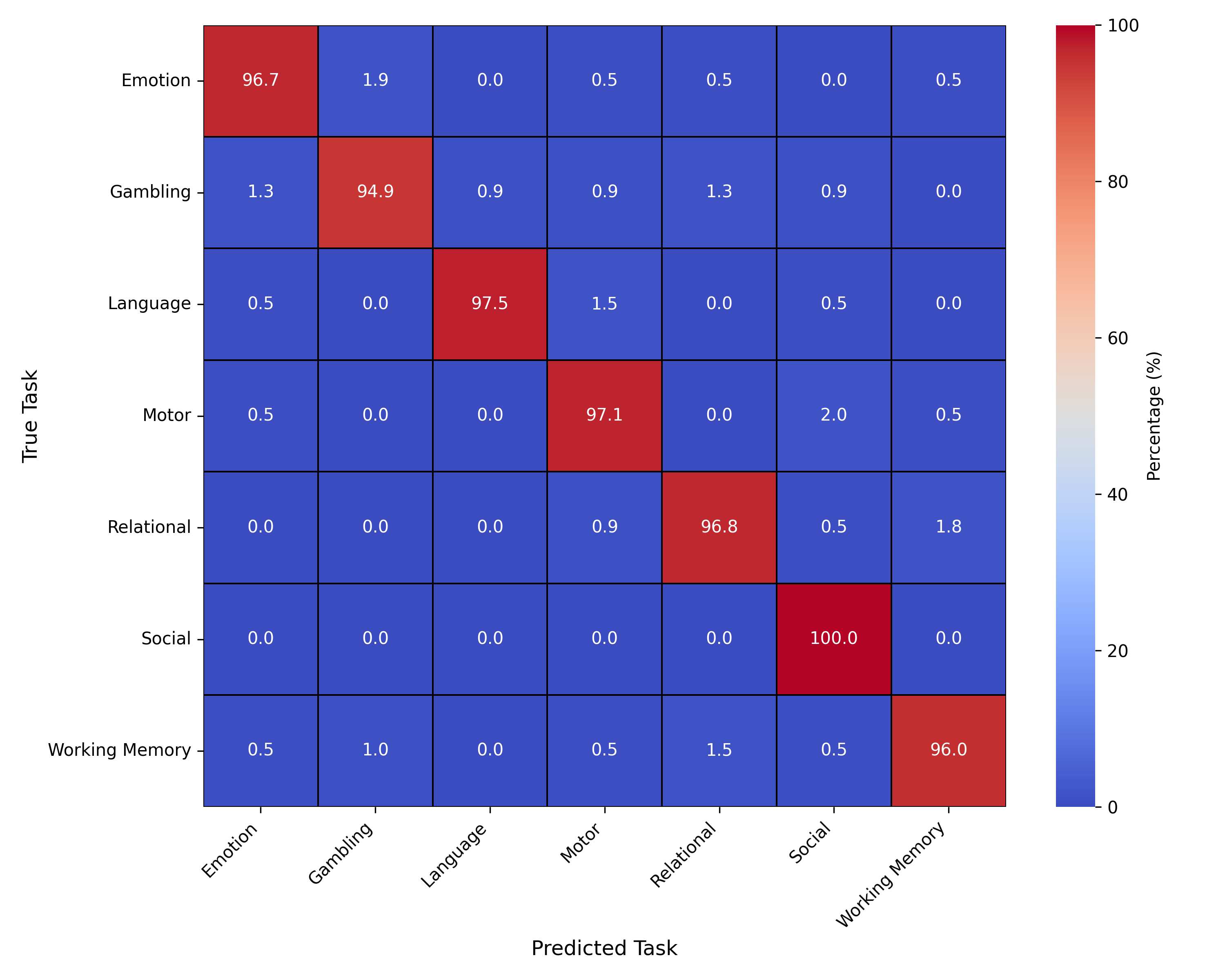}
    \caption{Confusion matrix of HCPTask classification task}
    \label{fig:confusion}
\end{figure}
\par The confusion matrix for the HCPTask classification task, shown in Fig.~\ref{fig:confusion}, indicates strong overall performance, with diagonal dominance reflecting accurate predictions across the seven cognitive paradigms i.e. Emotion, Gambling, Motor, Language, Relational, Social, and Working Memory. Out of 1,489 total samples, 1,444 are correctly classified, corresponding to an overall accuracy of approximately 97\%. This high accuracy suggests that the model effectively distinguishes among task states by capturing task-specific functional connectivity patterns in fMRI-derived brain graphs.

Minor confusions are evident in off-diagonal elements, such as 4 instances of true Emotion misclassified as Gambling, and 3 Gambling as Emotion, which may reflect overlapping neural circuits in reward and emotional processing (e.g., involving amygdala and ventral striatum). Similarly, Motor tasks show 4 misclassifications to Social, potentially due to shared action-observation networks in mirror neuron systems; Relational tasks have 4 to Working Memory and 2 to Motor, aligning with common frontoparietal involvement in executive functions. Language (3 to Motor) and Working Memory (3 to Relational) exhibit low errors, while Social is nearly perfect (209/209 correct), indicating robust separation for theory-of-mind processes.
\par Minor misclassifications appear in the off-diagonal entries and are limited in number. For example, four Emotion samples are misclassified as Gambling and three Gambling samples as Emotion, which may reflect shared neural substrates involved in affective and reward processing, such as the amygdala and ventral striatum\cite{santoro2024higher}. Motor tasks exhibit four misclassifications as Social, potentially due to overlap in action-observation and mirror neuron networks\cite{yang2015integrative}. Similarly, Relational tasks show a small number of confusions with Working Memory (four cases) and Motor tasks (two cases), consistent with the involvement of common frontoparietal circuits supporting executive functions\cite{barch2013function,cai2024multi}. Errors in Language (three misclassified as Motor) and Working Memory (three misclassified as Relational) remain minimal, while the Social task is classified perfectly (209/209), indicating strong separability of theory-of-mind-related activity\cite{schurz2020cross}.
\par Overall, these patterns highlight the model’s sensitivity to task-specific brain dynamics, with errors confined to biologically plausible overlaps between cognitive processes. This behavior supports the model’s applicability to neuroimaging classification tasks and motivates future analyses of region-wise contributions to further enhance interpretability.
\section{Conclusion}
In conclusion, SpectralBrainGNN represents a significant advancement in spectral graph neural networks for neuroimaging applications, achieving state of the art performance on HCP-derived benchmarks. In particular, it attains an accuracy of 96.25\% on the HCPTask dataset, outperforming strong baselines such as BrainMAP by 1.51\%. By using exact graph Fourier transforms with learnable frequency filters, the model captures multi-scale brain connectivity patterns, including task-specific activation pathways. The attention-based readout further improves interpretability by through sparse and biologically meaningful regions of interests.

\par Analysis of the confusion matrix confirms robust discrimination among cognitive states, with only minor misclassifications that align with known overlaps in neural circuitry, such as between emotion and gambling tasks. While the reliance on Laplacian eigendecomposition introduces scalability challenges for very large graphs, the model remains efficient for moderate sized connectomes through eigenvector caching strategies. Together with its strong generalization capability, these properties position SpectralBrainGNN as a promising framework for clinical applications, including brain disorder diagnosis. Future work will explore extensions to dynamic fMRI analysis and multimodal data fusion to further broaden its neuroscientific impact.


\begin{thebibliography}{10}
\providecommand{\url}[1]{\texttt{#1}}
\providecommand{\urlprefix}{URL }
\providecommand{\doi}[1]{https://doi.org/#1}

\bibitem{barch2013function}
Barch, D.M., Burgess, G.C., Harms, M.P., Petersen, S.E., Schlaggar, B.L., Corbetta, M., Glasser, M.F., Curtiss, S., Dixit, S., Feldt, C., et~al.: Function in the human connectome: task-fmri and individual differences in behavior. Neuroimage  \textbf{80},  169--189 (2013)

\bibitem{bassett2017network}
Bassett, D.S., Sporns, O.: Network neuroscience. Nature neuroscience  \textbf{20}(3),  353--364 (2017)

\bibitem{cai2024multi}
Cai, W., Taghia, J., Menon, V.: A multi-demand operating system underlying diverse cognitive tasks. Nature Communications  \textbf{15}(1), ~2185 (2024)

\bibitem{chen2024bridging}
Chen, J., Qi, Y., Wang, Y., Pan, G.: Bridging the semantic latent space between brain and machine: Similarity is all you need. In: Proceedings of the AAAI conference on artificial intelligence. vol.~38, pp. 11302--11310 (2024)

\bibitem{craik2019deep}
Craik, A., He, Y., Contreras-Vidal, J.L.: Deep learning for electroencephalogram (eeg) classification tasks: a review. Journal of neural engineering  \textbf{16}(3),  031001 (2019)

\bibitem{cui2022braingb}
Cui, H., Dai, W., Zhu, Y., Kan, X., Gu, A.A.C., Lukemire, J., Zhan, L., He, L., Guo, Y., Yang, C.: Braingb: a benchmark for brain network analysis with graph neural networks. IEEE transactions on medical imaging  \textbf{42}(2),  493--506 (2022)

\bibitem{cui2022positional}
Cui, H., Lu, Z., Li, P., Yang, C.: On positional and structural node features for graph neural networks on non-attributed graphs. In: Proceedings of the 31st ACM International Conference on Information \& Knowledge Management. pp. 3898--3902 (2022)

\bibitem{davis2020discovery}
Davis, K.D., Aghaeepour, N., Ahn, A.H., Angst, M.S., Borsook, D., Brenton, A., Burczynski, M.E., Crean, C., Edwards, R., Gaudilliere, B., et~al.: Discovery and validation of biomarkers to aid the development of safe and effective pain therapeutics: challenges and opportunities. Nature Reviews Neurology  \textbf{16}(7),  381--400 (2020)

\bibitem{eslami2019asd}
Eslami, T., Mirjalili, V., Fong, A., Laird, A.R., Saeed, F.: Asd-diagnet: a hybrid learning approach for detection of autism spectrum disorder using fmri data. Frontiers in neuroinformatics  \textbf{13}, ~70 (2019)

\bibitem{fox2007spontaneous}
Fox, M.D., Raichle, M.E.: Spontaneous fluctuations in brain activity observed with functional magnetic resonance imaging. Nature reviews neuroscience  \textbf{8}(9),  700--711 (2007)

\bibitem{gorban2021dynamic}
Gorban, A., Tyukina, T., Pokidysheva, L., Smirnova, E.: Dynamic and thermodynamic models of adaptation. Physics of life reviews  \textbf{37},  17--64 (2021)

\bibitem{hamilton2017inductive}
Hamilton, W., Ying, Z., Leskovec, J.: Inductive representation learning on large graphs. Advances in neural information processing systems  \textbf{30} (2017)

\bibitem{he2020deep}
He, T., Kong, R., Holmes, A.J., Nguyen, M., Sabuncu, M.R., Eickhoff, S.B., Bzdok, D., Feng, J., Yeo, B.T.: Deep neural networks and kernel regression achieve comparable accuracies for functional connectivity prediction of behavior and demographics. NeuroImage  \textbf{206},  116276 (2020)

\bibitem{jo2019deep}
Jo, T., Nho, K., Saykin, A.J.: Deep learning in alzheimer's disease: diagnostic classification and prognostic prediction using neuroimaging data. Frontiers in aging neuroscience  \textbf{11}, ~220 (2019)

\bibitem{kan2022brain}
Kan, X., Dai, W., Cui, H., Zhang, Z., Guo, Y., Yang, C.: Brain network transformer. Advances in Neural Information Processing Systems  \textbf{35},  25586--25599 (2022)

\bibitem{kawahara2017brainnetcnn}
Kawahara, J., Brown, C.J., Miller, S.P., Booth, B.G., Chau, V., Grunau, R.E., Zwicker, J.G., Hamarneh, G.: Brainnetcnn: Convolutional neural networks for brain networks; towards predicting neurodevelopment. NeuroImage  \textbf{146},  1038--1049 (2017)

\bibitem{keller2018task}
Keller, M., Taube, W., Lauber, B.: Task-dependent activation of distinct fast and slow (er) motor pathways during motor imagery. Brain stimulation  \textbf{11}(4),  782--788 (2018)

\bibitem{kipf2016semi}
Kipf, T.: Semi-supervised classification with graph convolutional networks. arXiv preprint arXiv:1609.02907  (2016)

\bibitem{kohoutova2020toward}
Kohoutov{\'a}, L., Heo, J., Cha, S., Lee, S., Moon, T., Wager, T.D., Woo, C.W.: Toward a unified framework for interpreting machine-learning models in neuroimaging. Nature protocols  \textbf{15}(4),  1399--1435 (2020)

\bibitem{li2021braingnn}
Li, X., Zhou, Y., Dvornek, N., Zhang, M., Gao, S., Zhuang, J., Scheinost, D., Staib, L.H., Ventola, P., Duncan, J.S.: Braingnn: Interpretable brain graph neural network for fmri analysis. Medical Image Analysis  \textbf{74},  102233 (2021)

\bibitem{li2022brain}
Li, Y., Zhang, X., Nie, J., Zhang, G., Fang, R., Xu, X., Wu, Z., Hu, D., Wang, L., Zhang, H., et~al.: Brain connectivity based graph convolutional networks and its application to infant age prediction. IEEE transactions on medical imaging  \textbf{41}(10),  2764--2776 (2022)

\bibitem{liao2024joint}
Liao, M., Wan, G., Du, B.: Joint learning neuronal skeleton and brain circuit topology with permutation invariant encoders for neuron classification. In: Proceedings of the AAAI Conference on Artificial Intelligence. vol.~38, pp. 197--205 (2024)

\bibitem{liu2023coupling}
Liu, X., Zhou, M., Shi, G., Du, Y., Zhao, L., Wu, Z., Liu, D., Liu, T., Hu, X.: Coupling artificial neurons in bert and biological neurons in the human brain. In: Proceedings of the AAAI Conference on Artificial Intelligence. vol.~37, pp. 8888--8896 (2023)

\bibitem{mohammed2023comprehensive}
Mohammed, A., Kora, R.: A comprehensive review on ensemble deep learning: Opportunities and challenges. Journal of King Saud University-Computer and Information Sciences  \textbf{35}(2),  757--774 (2023)

\bibitem{morris2019weisfeiler}
Morris, C., Ritzert, M., Fey, M., Hamilton, W.L., Lenssen, J.E., Rattan, G., Grohe, M.: Weisfeiler and leman go neural: Higher-order graph neural networks. In: Proceedings of the AAAI conference on artificial intelligence. vol.~33, pp. 4602--4609 (2019)

\bibitem{pei2022resgcn}
Pei, Y., Huang, T., Van~Ipenburg, W., Pechenizkiy, M.: Resgcn: attention-based deep residual modeling for anomaly detection on attributed networks. Machine Learning  \textbf{111}(2),  519--541 (2022)

\bibitem{rampavsek2022recipe}
Ramp{\'a}{\v{s}}ek, L., Galkin, M., Dwivedi, V.P., Luu, A.T., Wolf, G., Beaini, D.: Recipe for a general, powerful, scalable graph transformer. Advances in Neural Information Processing Systems  \textbf{35},  14501--14515 (2022)

\bibitem{richiardi2011decoding}
Richiardi, J., Eryilmaz, H., Schwartz, S., Vuilleumier, P., Van De~Ville, D.: Decoding brain states from fmri connectivity graphs. Neuroimage  \textbf{56}(2),  616--626 (2011)

\bibitem{saeidi2022decoding}
Saeidi, M., Karwowski, W., Farahani, F.V., Fiok, K., Hancock, P., Sawyer, B.D., Christov-Moore, L., Douglas, P.K.: Decoding task-based fmri data with graph neural networks, considering individual differences. Brain Sciences  \textbf{12}(8), ~1094 (2022)

\bibitem{said2023neurograph}
Said, A., Bayrak, R., Derr, T., Shabbir, M., Moyer, D., Chang, C., Koutsoukos, X.: Neurograph: Benchmarks for graph machine learning in brain connectomics. Advances in Neural Information Processing Systems  \textbf{36},  6509--6531 (2023)

\bibitem{sankar2018dynamic}
Sankar, A., Wu, Y., Gou, L., Zhang, W., Yang, H.: Dynamic graph representation learning via self-attention networks. arXiv preprint arXiv:1812.09430  (2018)

\bibitem{santoro2024higher}
Santoro, A., Battiston, F., Lucas, M., Petri, G., Amico, E.: Higher-order connectomics of human brain function reveals local topological signatures of task decoding, individual identification, and behavior. Nature Communications  \textbf{15}(1),  10244 (2024)

\bibitem{saranskaia2025aim}
Saranskaia, I., Gutkin, B., Zakharov, D.: Aim-based choice of strategy for meg-based brain state classification. The European Physical Journal Special Topics pp. 1--19 (2025)

\bibitem{schurz2020cross}
Schurz, M., Maliske, L., Kanske, P.: Cross-network interactions in social cognition: a review of findings on task related brain activation and connectivity. cortex  \textbf{130},  142--157 (2020)

\bibitem{takerkart2014graph}
Takerkart, S., Auzias, G., Thirion, B., Ralaivola, L.: Graph-based inter-subject pattern analysis of fmri data. PloS one  \textbf{9}(8),  e104586 (2014)

\bibitem{vaghari2022late}
Vaghari, D., Kabir, E., Henson, R.N.: Late combination shows that meg adds to mri in classifying mci versus controls. Neuroimage  \textbf{252},  119054 (2022)

\bibitem{van2013wu}
Van~Essen, D.C., Smith, S.M., Barch, D.M., Behrens, T.E., Yacoub, E., Ugurbil, K., Consortium, W.M.H., et~al.: The wu-minn human connectome project: an overview. Neuroimage  \textbf{80},  62--79 (2013)

\bibitem{velivckovic2017graph}
Veli{\v{c}}kovi{\'c}, P., Cucurull, G., Casanova, A., Romero, A., Lio, P., Bengio, Y.: Graph attention networks. arXiv preprint arXiv:1710.10903  (2017)

\bibitem{vlasenko2024ensemble}
Vlasenko, D., Zaikin, A., Zakharov, D.: Ensemble methods for representation of fmri, eeg/meg data in graph form for classification of brain states. In: 2024 8th Scientific School Dynamics of Complex Networks and their Applications (DCNA). pp. 258--261. IEEE (2024)

\bibitem{wang2024graph}
Wang, C., Tsepa, O., Ma, J., Wang, B.: Graph-mamba: Towards long-range graph sequence modeling with selective state spaces. arXiv preprint arXiv:2402.00789  (2024)

\bibitem{wang2010graph}
Wang, J., Zuo, X., He, Y.: Graph-based network analysis of resting-state functional mri. Frontiers in systems neuroscience  \textbf{4}, ~1419 (2010)

\bibitem{wang2022graph}
Wang, S., Chen, C., Li, J.: Graph few-shot learning with task-specific structures. Advances in Neural Information Processing Systems  \textbf{35},  38925--38936 (2022)

\bibitem{wang2025brainmap}
Wang, S., Lei, Z., Tan, Z., Ding, J., Zhao, X., Dong, Y., Wu, G., Chen, T., Chen, C., Zhang, A., et~al.: Brainmap: Learning multiple activation pathways in brain networks. In: Proceedings of the AAAI Conference on Artificial Intelligence. vol.~39, pp. 14432--14440 (2025)

\bibitem{xu2024learning}
Xu, Y., Peng, Z., Shi, B., Hua, X., Dong, B.: Learning dynamic graph representations through timespan view contrasts. Neural Networks  \textbf{176},  106384 (2024)

\bibitem{yang2015integrative}
Yang, D.Y.J., Rosenblau, G., Keifer, C., Pelphrey, K.A.: An integrative neural model of social perception, action observation, and theory of mind. Neuroscience \& Biobehavioral Reviews  \textbf{51},  263--275 (2015)

\bibitem{yang2022data}
Yang, Y., Zhu, Y., Cui, H., Kan, X., He, L., Guo, Y., Yang, C.: Data-efficient brain connectome analysis via multi-task meta-learning. In: Proceedings of the 28th ACM SIGKDD Conference on Knowledge Discovery and Data Mining. pp. 4743--4751 (2022)

\bibitem{zhang2022probing}
Zhang, X., Wang, S., Lin, N., Zhang, J., Zong, C.: Probing word syntactic representations in the brain by a feature elimination method. In: Proceedings of the AAAI Conference on Artificial Intelligence. vol.~36, pp. 11721--11729 (2022)

\bibitem{zhang2024metarlec}
Zhang, Z., Ji, J., Liu, J.: Metarlec: Meta-reinforcement learning for discovery of brain effective connectivity. In: Proceedings of the AAAI Conference on Artificial Intelligence. vol.~38, pp. 10261--10269 (2024)

\end{thebibliography}
\end{document}